\documentclass[sigconf]{acmart}

\usepackage{float}
\usepackage{multirow}
\usepackage{booktabs}

\AtBeginDocument{%
  }

\setcopyright{acmlicensed}
\copyrightyear{2018}
\acmYear{2018}
\acmDOI{XXXXXXX.XXXXXXX}
\acmConference[Conference acronym 'XX]{Make sure to enter the correct
  conference title from your rights confirmation email}{June 03--05,
  2018}{Woodstock, NY}
\acmISBN{978-1-4503-XXXX-X/2018/06}

\begin{document}

\title{The First MPDD Challenge: Multimodal Personality-aware Depression Detection}


%
\author{Changzeng Fu}
\affiliation{%
  \institution{Northeastern University}
  \city{Shenyang}
  \country{China}}
\email{fuchangzeng@qhd.neu.edu.cn}


\author{Zelin Fu}
\affiliation{%
  \institution{Northeastern University}
  \city{Shenyang}
  \country{China}}
\email{202219117@stu.neu.edu.cn}

\author{Qi Zhang}
\affiliation{%
  \institution{Northeastern University}
  \city{Shenyang}
  \country{China}}
\email{202212078@stu.neu.edu.cn}

\author{Xinhe Kuang}
\affiliation{%
  \institution{Northeastern University}
  \city{Shenyang}
  \country{China}}
\email{202319257@stu.neuq.edu.cn}

\author{Jiacheng Dong}
\affiliation{%
  \institution{University of Technology Sydney}
  \city{Sydney}
  \country{Australia}}
\email{jiacheng.dong@student.uts.edu.au}

\author{Kaifeng Su}
\affiliation{%
  \institution{Northeastern University}
  \city{Shenyang}
  \country{China}}
\email{2372306@stu.neu.edu.cn}

\author{Yikai Su}
\affiliation{%
  \institution{Northeastern University}
  \city{Shenyang}
  \country{China}}
\email{2372307@stu.neu.edu.cn}


\author{Wenbo Shi}
\affiliation{%
  \institution{Northeastern University}
  \city{Shenyang}
  \country{China}}
\email{shiwenbo@neuq.edu.cn}

\author{Junfeng Yao}
\affiliation{%
  \institution{Xiamen University}
  \city{Xiamen}
  \country{China}}
\email{yao0010@xmu.edu.cn}

\author{Yuliang Zhao}
\affiliation{%
  \institution{Northeastern University}
  \city{Shenyang}
  \country{China}}
\email{zhaoyuliang@neuq.edu.cn}

\author{Shiqi Zhao}
\affiliation{%
  \institution{Northeastern University}
  \city{Shenyang}
  \country{China}}
\email{zhaoshiqi@qhd.neu.edu.cn}

\author{Jiadong Wang}
\affiliation{%
  \institution{Technical University of Munich}
  \city{Munich}
  \country{Germany}}
\email{jiadong.wang@tum.de}

\author{Siyang Song}
\affiliation{%
  \institution{University of Cambridge}
  \city{Cambridge}
  \country{UK}}
\email{ss2796@cam.ac.uk}

\author{Chaoran Liu}
\affiliation{%
  \institution{National Information Institute}
  \city{Tokyo}
  \country{Japan}}
\email{chaoran.liu@riken.jp}

\author{Yuichiro Yoshikawa}
\affiliation{%
  \institution{Osaka University}
  \city{Osaka}
  \country{Japan}}
\email{yoshikawa@irl.sys.es.osaka-u.ac.jp}

\author{Björn Schuller}
\affiliation{%
  \institution{Imperial College London}
  \city{London}
  \country{UK}}
\email{bjoern.schuller@imperial.ac.uk}

\author{Hiroshi Ishiguro}
\affiliation{%
  \institution{Osaka University}
  \city{Osaka}
  \country{Japan}}
\email{ishiguro@irl.sys.es.osaka-u.ac.jp}







\renewcommand{\shortauthors}{Fu et al.}

\begin{abstract}
Depression is a widespread mental health issue affecting diverse age groups, with notable prevalence among college students and the elderly. However, existing datasets and detection methods primarily focus on young adults, neglecting the broader age spectrum and individual differences that influence depression manifestation. Current approaches often establish a direct mapping between multimodal data and depression indicators, failing to capture the complexity and diversity of depression across individuals. This challenge includes two tracks based on age-specific subsets: \textbf{Track 1} uses the MPDD-Elderly dataset for detecting depression in older adults, and \textbf{Track 2} uses the MPDD-Young dataset for detecting depression in younger participants. The Multimodal Personality-aware Depression Detection (MPDD) Challenge aims to address this gap by incorporating multimodal data alongside individual difference factors. We provide a baseline model that fuses audio and video modalities with individual difference information to detect depression manifestations in diverse populations. This challenge aims to promote the development of more personalized and accurate de pression detection methods, advancing mental health research and fostering inclusive detection systems. More details are available on the official challenge website: \textcolor{blue}{\url{https://hacilab.github.io/MPDDChallenge.github.io}}.
\end{abstract}



\received{20 February 2007}
\received[revised]{12 March 2009}
\received[accepted]{5 June 2009}

\maketitle

\section{Introduction}
Depression is a pervasive mental health disorder with a profound impact on global well-being, affecting millions and significantly hindering the quality of life, productivity, and overall health \cite{moussavi2007depression}. The World Health Organization (WHO) categorizes depression as a leading cause of disability, underscoring its substantial societal and economic burden \cite{world2017depression}. Timely detection and intervention are pivotal in mitigating the chronic and debilitating effects of depression. However, traditional methods, which rely primarily on self-reported questionnaires such as PHQ-9 and BDI-II, are limited in their capacity to capture the dynamic and multifaceted nature of depressive symptoms. These methods are also prone to reporting biases and may fail to detect early or subtle changes in depressive states.

The advent of specialized datasets has been instrumental in advancing depression research by providing multimodal data that can enhance the detection and understanding of depressive symptoms. For instance, the AVEC challenge has contributed rich audiovisual interview data annotated with depression-related information, facilitating more nuanced assessments \cite{valstar2013avec, valstar2014avec, valstar2016avec, ringeval2017avec, ringeval2019avec}. The DAIC-WoZ dataset has furthered our understanding by incorporating clinical interviews and PHQ-9 scores \cite{gratch2014distress}. Other datasets, including the Pittsburgh dataset \cite{dibekliouglu2017dynamic}, the D-Vlog dataset \cite{yoon2022d}, the EATD-Corpus \cite{shen2022automatic}, the CMDC \cite{zou2022semi}, and the MODMA \cite{cai2022multi}, have expanded the scope by integrating diverse data types like EEG and utilizing standard depression scales for evaluation. Additionally, datasets such as AMIGOS \cite{miranda2018amigos} and DEAP \cite{koelstra2011deap} have incorporated physiological data, offering insights into the physiological underpinnings of depression, while the SEED dataset provides EEG and eye movement data annotated with personality traits \cite{zheng2015investigating}, highlighting the role of individual differences in depressive symptoms.


Despite the burgeoning interest in multimodal approaches for depression detection, existing datasets and methodologies exhibit notable limitations. Many datasets primarily focus on establishing direct correlations between behavioral signals and depression levels. This approach, while valuable, often overlooks the nuanced individual variations in depressive symptoms \cite{klein2011personality,lo2017genome}. Moreover, the demographic scope of existing datasets is predominantly confined to young adults, thereby overlooking other critical age groups, such as the elderly, who may present distinct patterns of depression.

\begin{table}[htbp]
    \centering
    \caption{Comparison of Demographic Statistics Between MPDD-Young and MPDD-Elderly}
    \label{tab:dataset_comparison}
    \renewcommand{\arraystretch}{1.2}
    \setlength{\tabcolsep}{8pt}
    \begin{tabular}{lccc}
        \toprule
        \textbf{Dataset} & \textbf{Avg. Age (Years)} & \textbf{Male (\%)} & \textbf{Female (\%)} \\
        \midrule
        MPDD-Young & 20.05 $\pm$ 2.23 & 51.82\% & 48.18\% \\
        MPDD-Elderly & 62.78 $\pm$ 11.02 & 50.00\% & 50.00\% \\
        \bottomrule
    \end{tabular}
\end{table}

\section{Dataset}

\begin{table*}[h]
    \centering
    \caption{Comparison of Datasets for Depression Detection}
    \begin{tabular}{lccccccc}
        \hline
        Dataset  & Multimodal & Depression & Personality & Gender & Age & Region & Disease \\
        \hline
        AVEC &  \checkmark & \checkmark & - & \checkmark & \checkmark & - & - \\
        DAIC-WoZ & \checkmark & \checkmark & - & \checkmark & - & - & -\\
        Pittsburgh & \checkmark & \checkmark & - & \checkmark & \checkmark & - & - \\
        D-Vlog & \checkmark & \checkmark & - & \checkmark & - & - & - \\
        MMDA  & \checkmark& \checkmark & - & \checkmark & \checkmark & - & - \\
        EATD-Corpus & \checkmark& \checkmark & - & - & - & - & - \\
        CMDC  & \checkmark& \checkmark & - & \checkmark & \checkmark & - & - \\
        MODMA & \checkmark & \checkmark & - & \checkmark & \checkmark & - & - \\
        \hline
        MPDD (ours) & \checkmark & \checkmark & \checkmark & \checkmark & \checkmark & \checkmark & \checkmark \\
        \hline
    \end{tabular}
    \label{tab:datasets}
\end{table*}

\subsection{Novelty}

\begin{figure*}
    \centering
    \includegraphics[width=0.85\linewidth]{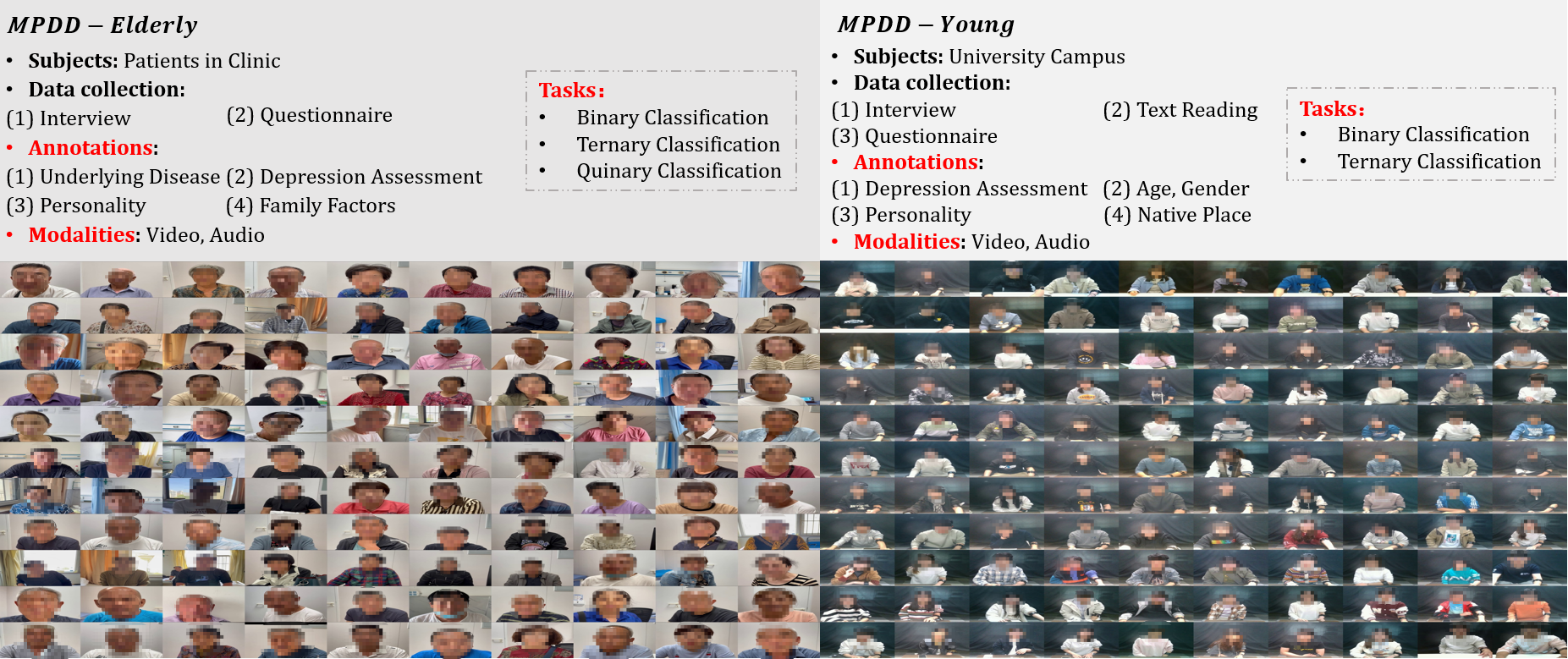}
    \caption{Overview of the MPDD Dataset. }
    \label{fig:ds_overview}
\end{figure*}

To address these limitations, we introduce the \textit{First MPDD Challenge: Multimodal Personality-aware Depression Detection}, complemented by a novel dataset. This initiative presents a multimodal dataset that incorporates audio and visual data from participants engaged in a variety of real-life scenarios. The MPDD dataset is enriched with annotations grounded in the PHQ-9 scale, the Big Five Personality traits, and an array of demographic details.

The dataset consists of two subsets, \textbf{MPDD-Young} and \textbf{MPDD-Elderly}, as shown in figure~\ref{fig:ds_overview}, focusing on depression detection among young individuals and elderly participants, respectively. Each subset is independently collected and annotated, enabling targeted analysis of age-specific patterns in multimodal depressive expression.
Compared to existing datasets, the MPDD dataset significantly enhances contextual diversity and annotation depth, as highlighted in Table~\ref{tab:datasets}. Its comprehensive annotations capture individual differences, providing a more inclusive and granular depiction of depression. 

\subsection{MPDD-Elderly}


The MPDD-Elderly dataset includes data from elderly participants collected at a Hospital in China, focusing on elderly individuals with underlying medical conditions. Before participation, each volunteer received an informed consent form and could choose which parts of the experiment to participate in. The gender distribution and average age of the dataset are summarized in Table~\ref{tab:dataset_comparison}.
%

A semi-structured interview (SSI) approach was used to systematically assess the psychological and emotional state of participants. SSI combines both closed and open-ended questions, often followed by “why” or “how” inquiries to encourage deeper responses. Instead of strictly adhering to a standardized script, the conversation is guided by predefined topics, allowing for more flexibility. The interview questions were designed based on depression assessment scales such as HAMD and BDI, with special consideration for elderly individuals with chronic illnesses.
During the interview, participants were encouraged to answer questions freely, and appropriate prompts were given to participants who were less expressive to ensure adequate acquisition of information. The entire session was recorded using a portable camera, capturing both facial expressions and audio.
During data collection, participants completed the PHQ-9 and HAMD-24 scales to assess depression severity, and the Big Five Personality-10 (BigFive-10) scale to evaluate personality traits.

The depression labels are derived from these scales: HAMD-24 supports binary and ternary classification tasks, while PHQ-9 is used for the quinary classification task. 
Based on these annotations, the dataset supports the following three classification tasks:

\begin{itemize}
        \item Binary classification (normal / depressed)
        \item Ternary classification (normal / mildly depressed / severely depressed)
        \item Quinary classification (normal / mildly / moderate depressed / moderately severe / severe)
\end{itemize}


To support more detailed characterization of each participant, the MPDD-Elderly dataset includes additional annotations that describe personal and contextual information. These labels cover physical health conditions (e.g., endocrine or neurological diseases), financial stress levels, and the number of co-habiting family members. Personality traits from the BigFive-10 scale are also included to reflect individual behavioral tendencies. These annotations help build a more complete profile for each participant, which can support personalized analysis. A summary of all label definitions is shown in Table~\ref{tab:elderly_label_definitions}.

\begin{table}[h]
\renewcommand{\arraystretch}{1.2}
\centering
\caption{MPDD-Young Personalized Label Definitions}
\begin{tabular}{p{2.5cm}p{5cm}}  
\toprule
\textbf{Label} & \textbf{Description} \\
\midrule
\textbf{Big5 traits} & One-hot style encoding of the top-scoring BigFive traits (ties allowed) \\
\textbf{Disease category} & 0 = healthy, 1 = other, 2 = endocrine, 3 = circulatory, 4 = neurological \\
\textbf{Financial stress} & Economic pressure: 0 = none, 1 = mild, 2 = moderate, 3 = severe/unbearable \\
\textbf{Family members} & Number of cohabiting individuals \\
\bottomrule
\end{tabular}
\label{tab:elderly_label_definitions}
\end{table}

\subsection{MPDD-Young}

The MPDD-Young dataset includes data from young adult participants, focusing on non-clinical populations experiencing varying levels of depressive symptoms. Before participation, each subject also provided written informed consent. The gender distribution and average age of the dataset are also summarized in Table~\ref{tab:dataset_comparison}.

The data collection process included three stages: questionnaire completion, a self-introduction session, and two controlled text reading tasks. Throughout the process, facial expressions were recorded via a front-facing camera, and speech audio was captured using a microphone. These stages were designed to capture diverse emotional behaviors under both spontaneous and guided conditions.
During data collection, participants completed both the PHQ-9 scale and the BigFive-10 scale to assess depression severity and personality traits. 

Depression labels are derived from PHQ-9 scores: binary classification labels distinguish between depressed and non-depressed individuals, while ternary classification labels categorize participants into normal, mildly depressed, and severely depressed groups. Based on these annotations, the dataset supports the following two classification tasks:

\begin{itemize}
\item Binary classification (normal / depressed)
\item Ternary classification (normal / mildly depressed / severely depressed)
\end{itemize}



In addition to depression-related annotations, the MPDD-Young dataset provides information on participant background to support personalized analysis. These include demographic variables such as age, gender, and place of origin, which enable comparisons across different population groups. The dataset also includes personality trait labels based on the BigFive-10 scale, presented in a binary form indicating the dominant personality characteristics. The definitions of all labels are summarized in Table~\ref{tab:young_label_definitions}.

\begin{table}[h]
\renewcommand{\arraystretch}{1.2}
\centering
\caption{MPDD-Young Personalized Label Definitions}
\begin{tabular}{p{2.5cm}p{5cm}}
\toprule
\textbf{Label} & \textbf{Description} \\
\midrule
\textbf{Age} & Age of the participant (in years) \\
\textbf{Gender} & Participant's gender \\
\textbf{Native place} & Birthplace or region of origin  \\
\textbf{Big5 traits} & One-hot style encoding of the top-scoring BigFive traits (ties allowed) \\
\bottomrule
\end{tabular}
\label{tab:young_label_definitions}
\end{table}

\section{Methods}

\subsection{Baseline System}

\begin{figure*}[t]
\centering
\includegraphics[width=0.7\linewidth]{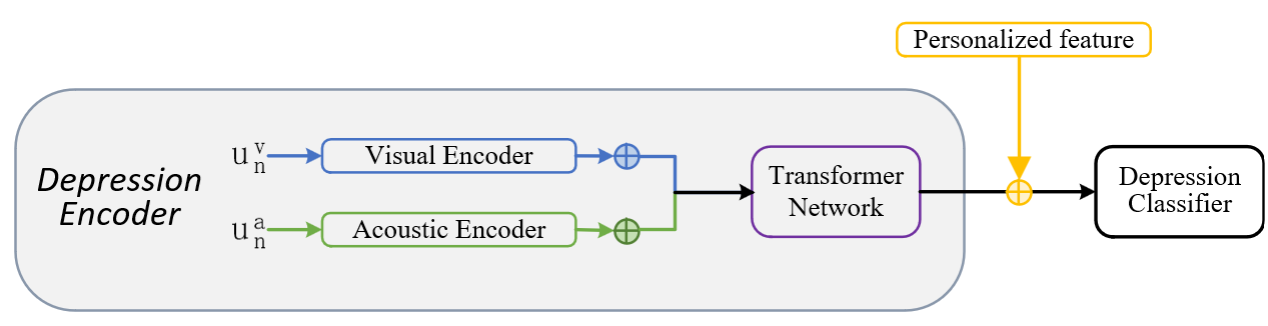}
\vspace{-1em}
\caption{The Overall Framework for Baseline Model.}
   \label{baseline model}
\end{figure*}

The baseline model for the MPDD Challenge is designed as a multimodal depression encoder that integrates visual and acoustic features to predict depression levels. The model architecture, as depicted in Figure~\ref{baseline model}, consists of the following main components:

\paragraph{Visual Encoder} The visual encoder is implemented using an LSTM-based feature extractor. It processes sequential visual input to capture facial expressions and movement patterns associated with depression. 

\paragraph{Acoustic Encoder} The acoustic encoder also employs an LSTM-based architecture, designed to process speech input. It captures prosodic and vocal characteristics such as tone, pitch, and rhythm.
    
\paragraph{Transformer Fusion Network} The outputs from the visual and acoustic encoders are fused by concatenating. This fusion allows the model to combine the complementary information from both modalities, leading to a more robust representation of the input data. The fused features are then fed into a transformer network. This component helps to capture contextual dependencies within the multimodal data.

\paragraph{Indivisual Embedding} To incorporate personalized factors, after learning temporal dependencies through the transformer network, a feature embedding representing personalized factors is concatenated to the multimodal representation. 
    
\paragraph{Depression Classifier} After the Transformer Fusion Network outputs the fused features along with personalized embeddings, these features are fed into the depression classifier, which predicts the depression level based on the encoded representations.

The model architecture enables both multimodal feature learning and personalized feature learning, addressing previous research limitations in capturing individual differences in depression detection, and provides a fundamental framework for the MPDD Challenge.

\subsection{Feature Sets}

In the MPDD Challenge, diverse audio and video features are extracted to enhance depression detection. These features include emotional expressions and physiological responses linked to depressive symptoms. The feature sets are extracted over two distinct time windows: 1-second and 5-second intervals, allowing the model to analyze both short-term and longer-term emotional states.

\paragraph{Audio Features}
Audio features play a crucial role in detecting depression as they capture speech patterns, tone, and other vocal characteristics. To comprehensively compare the impact of different feature extraction methods on classification performance, we employ the following audio features:

\begin{itemize}
    \item \textbf{MFCC (Mel-frequency cepstral coefficients)}\footnote{https://github.com/librosa/librosa}: A widely used feature in speech processing that represents the short-term power spectrum of a sound.
    \item \textbf{OpenSMILE}\footnote{https://github.com/audeering/opensmile}: An open-source software toolkit for audio feature extraction, which provides a comprehensive set of low-level descriptors capturing various acoustic properties. Specifically, we adopted the ComParE 2016 configuration.
    \item \textbf{Wav2Vec}\footnote{https://github.com/facebookresearch/fairseq/tree/main/examples/wav2vec}: A deep learning model pre-trained on large amounts of unlabeled audio data, capable of extracting meaningful representations from raw audio waveforms.
\end{itemize}

\paragraph{Video Features}
Video features are essential for analyzing facial expressions and other visual cues associated with depression. To compare the impact of different feature extraction methods on classification performance, we employ the following video features:  

\begin{itemize}
    \item \textbf{DenseNet (Densely Connected Convolutional Networks)}\footnote{https://github.com/liuzhuang13/DenseNet}: A deep learning model known for its effectiveness in image classification tasks, used here to extract features from facial images.
    \item \textbf{OpenFace}\footnote{http://multicomp.cs.cmu.edu/resources/openface/}: A facial behavior analysis toolkit that provides a set of features related to facial expressions and head pose.
    \item \textbf{ResNet (Residual Networks)}\footnote{https://huggingface.co/microsoft/resnet-50}: A deep learning model that uses residual connections to ease the training of deep networks, employed for feature extraction from visual data.
\end{itemize}

The dataset is split into training and testing sets. Within the training set, 90\% of the data is used for training and the remaining 10\% serves as the validation set. 
The splitting is strictly based on subject IDs to ensure that samples from the same individual do not appear in both training and testing sets.
For the MPDD-Elderly subset, both the 1s and 5s sliding windows generate a total of 564 samples, with 337 used for training and 227 for testing.
For the MPDD-Young subset, each window setting produces 528 samples, evenly split into 264 training and 264 testing samples. A detailed breakdown is provided in Table~\ref{tab:sample_split}.

\begin{table}[h]
\centering
\small
\caption{Sample Counts in Dataset After Sliding Window}
\vspace{-1em}
\label{tab:sample_split}
\setlength{\tabcolsep}{4pt}
\renewcommand{\arraystretch}{1.2}
\begin{tabular}{lccc}
\toprule
\textbf{Subset} & \textbf{Total} & \textbf{Train} & \textbf{Test} \\
\midrule
MPDD-Elderly  & 564 & 337 & 227 \\
MPDD-Young    & 528 & 264 & 264 \\
\bottomrule
\end{tabular}
\end{table}
\vspace{-1em}

\subsection{Personalized Features }

To enhance personalization, we introduce individualized textual representations derived from participant-level attributes such as BigFive personality traits and financial stress levels. These attributes are inserted into a structured prompt and processed by a generative language model (ChatGLM3) to produce a concise English description of each participant’s emotional expression style. An example prompt includes factual statements about each trait followed by an instruction asking for a short behavioral summary without referencing depression-related content.

The generated descriptions are encoded using the RoBERTa-large model , where the sentence embedding is extracted from the final hidden state of the first token. These 1024-dimensional embeddings serve as compact personalized features and are fused with acoustic and visual features for downstream classification.

\subsection{Metrics}

We use a combination of weighted and unweighted metrics to evaluate performance in the MPDD Challenge.  
Accuracy measures the overall proportion of correct predictions:
\begin{equation}
\text{Acc} = \frac{\text{Number of Correct Predictions}}{\text{Total Number of Samples}}
\end{equation}

F1 score balances precision and recall, which is computed as:
\begin{equation}
F_1 = 2 \times \frac{\text{Precision} \times \text{Recall}}{\text{Precision} + \text{Recall}}
\end{equation}
where:
\begin{align*}
\text{Precision} &= \frac{TP}{TP + FP} \\
\text{Recall} &= \frac{TP}{TP + FN}
\end{align*}

For each task, we report both weighted and unweighted accuracy and F1 score. The final metric for each task is the average of weighted and unweighted values:
\begin{equation}
\begin{aligned}
Acc^{task} &= \frac{Acc_{weighted} + Acc_{unweighted}}{2} \\
F_1^{task} &= \frac{F_{1,weighted} + F_{1,unweighted}}{2}
\end{aligned}
\end{equation}

To assess the overall model performance for each track, we compute the mean of task-level scores:
\begin{equation}
\begin{aligned}
  Acc_{track} & = \text{Average}(Acc^{task1}, Acc^{task2}, \dots) \\
  F_{track} & = \text{Average}(F_1^{task1}, F_1^{task2}, \dots)
\end{aligned}
\end{equation}

\section{Results}

\subsection{Competitive Results}

\begin{table*}[h!]
    \centering
    \caption{Baseline Results of MPDD-Elderly (PF: personalized features; W/U: weighted/unweighted)}
    \label{baselineresult-MPDDE}
    \small
    \setlength{\tabcolsep}{4pt}
    \renewcommand{\arraystretch}{1.2}
    \begin{tabular}{@{}lccccccccccc@{}}
        \toprule
        \multirow{2}{*}{Length} & \multirow{2}{*}{Task} & \multirow{2}{*}{Audio} & \multirow{2}{*}{Visual} 
        & \multicolumn{4}{c}{w/o PF} & \multicolumn{4}{c}{w/ PF} \\
        \cmidrule(lr){5-8} \cmidrule(lr){9-12}
         & & & & W\_Acc & U\_Acc & W\_F1 & U\_F1 & W\_Acc & U\_Acc & W\_F1 & U\_F1 \\
        \midrule
        \multirow{3}{*}{1s}
            & Binary  & mfcc      & openface  & 69.37 & 83.33 & 82.60 & 70.89 & 85.40 & 84.62 & 85.71 & 79.13 \\
            & Ternary & opensmile & resnet    & 48.93 & 55.13 & 54.35 & 49.14 & 55.49 & 56.41 & 56.48 & 55.64 \\
            & Quinary & opensmile & densenet  & 42.45 & 66.67 & 63.85 & 44.00 & 45.79 & 69.23 & 66.26 & 46.66 \\
        \midrule
        \multirow{3}{*}{5s}
            & Binary  & opensmile & resnet    & 67.94 & 76.92 & 77.90 & 66.15 & 75.40 & 80.77 & 81.75 & 72.37 \\
            & Ternary & Wav2Vec   & openface  & 46.58 & 50.00 & 50.88 & 47.59 & 59.62 & 57.69 & 58.22 & 59.37 \\
            & Quinary & mfcc      & densenet  & 56.98 & 75.64 & 73.49 & 56.83 & 57.71 & 78.21 & 75.62 & 58.40 \\
        \bottomrule
    \end{tabular}
\end{table*}

\begin{table*}[h!]
    \centering
    \caption{Baseline Results of MPDD-Young (PF: personalized features; W/U: weighted/unweighted)}
    \label{baselineresult-MPDDY}
    \small
    \setlength{\tabcolsep}{6pt}
    \renewcommand{\arraystretch}{1.2}
    \begin{tabular}{@{}lccccccccccc@{}}
        \toprule
        \multirow{2}{*}{Length} & \multirow{2}{*}{Task} & \multirow{2}{*}{Audio} & \multirow{2}{*}{Visual} 
        & \multicolumn{4}{c}{w/o PF} & \multicolumn{4}{c}{w/ PF} \\
        \cmidrule(lr){5-8} \cmidrule(lr){9-12}
         & & & & W\_Acc & U\_Acc & W\_F1 & U\_F1 & W\_Acc & U\_Acc & W\_F1 & U\_F1 \\
        \midrule
        \multirow{2}{*}{1s}
            & Binary  & Wav2Vec    & openface  & 56.06 & 56.06 & 55.23 & 55.23 & 63.64 & 63.64 & 59.96 & 59.96 \\
            & Ternary & mfcc       & densenet  & 42.63 & 48.48 & 47.95 & 43.72 & 49.66 & 51.52 & 51.86 & 51.62 \\
        \midrule
        \multirow{2}{*}{5s}
            & Binary  & opensmile  & resnet    & 60.61 & 60.61 & 60.02 & 60.02 & 62.12 & 62.12 & 62.11 & 62.11 \\
            & Ternary & mfcc       & densenet  & 41.29 & 42.42 & 42.82 & 39.38 & 41.71 & 50.00 & 48.18 & 41.31 \\
        \bottomrule
    \end{tabular}
\end{table*}

The MPDD-Elderly dataset results show that in the binary classification task with 1-second clips, the model using MFCC audio features and OpenFace visual features achieved the highest weighted accuracy of 85.40\% and a weighted F1 score of 85.71\% when personalized features were included. In ternary classification with the same clip length, the best result was obtained using OpenSMILE audio features and ResNet visual features, reaching 55.49\% weighted accuracy and 56.48\% weighted F1. For quinary classification, the combination of OpenSMILE and DenseNet achieved 45.79\% weighted accuracy and 66.26\% weighted F1. 
With 5-second clips, the binary classification model using OpenSMILE and ResNet reached 75.40\% weighted accuracy and 81.75\% weighted F1. In ternary classification, Wav2Vec and OpenFace yielded the best performance with 59.62\% weighted accuracy and 58.22\% weighted F1. For the quinary task, the MFCC and DenseNet combination resulted in 57.71\% weighted accuracy and 75.62\% weighted F1 with personalized features.

The results for the MPDD-Young dataset indicate that in binary classification with 1-second clips, the model using Wav2Vec and OpenFace reached 63.64\% weighted accuracy and 59.96\% weighted F1. For ternary classification, MFCC and DenseNet achieved 49.66\% weighted accuracy and 51.86\% weighted F1. Using 5-second clips, the binary classification model based on OpenSMILE and ResNet reached 62.12\% weighted accuracy and 62.11\% weighted F1. For the ternary task, MFCC and DenseNet achieved 41.71\% weighted accuracy and 48.18\% weighted F1 when personalized features were included.


\subsection{Ablation Study}
To evaluate the effectiveness of incorporating personalized features (PF), we conduct an ablation study by comparing the performance of models with and without PF across different tasks and datasets.

As shown in Table~\ref{baselineresult-MPDDE}, adding PF consistently improves classification performance on the MPDD-Elderly dataset. For binary classification with 1-second segments, weighted accuracy increases from 69.37\% to 85.40\%, and weighted F1-score improves from 82.60 to 85.71. In the ternary task, PF boosts weighted accuracy from 48.93\% to 55.49\%, and F1 from 54.35 to 56.48. Similarly, in quinary classification, accuracy rises from 42.45\% to 45.79\%, and F1 from 63.85 to 66.26. For 5-second segments, the performance gain remains consistent: binary accuracy improves from 67.94\% to 75.40\%, and F1 from 77.90 to 81.75; ternary accuracy increases from 46.58\% to 59.62\%, and F1 from 50.88 to 58.22; quinary accuracy improves slightly from 56.98\% to 57.71\%, while F1 rises from 73.49 to 75.62.

A similar trend is observed in Table~\ref{baselineresult-MPDDY} for the MPDD-Young dataset. In binary classification with 1-second segments, the weighted accuracy increases from 56.06\% to 63.64\%, and F1-score from 55.23 to 59.96. For ternary classification, accuracy improves from 42.63\% to 49.66\%, and F1 from 47.95 to 51.86. In 5-second settings, binary classification shows a smaller but consistent gain, with accuracy rising from 60.61\% to 62.12\%, and F1 from 60.02 to 62.11. In the ternary task, accuracy improves from 41.29\% to 41.71\%, and F1 from 42.82 to 48.18. These results confirm that incorporating personalized features leads to performance improvements across different segment lengths and classification tasks.


\section{Conclusion}
This year’s MPDD Challenge introduces the theme of “Multimodal Personality-aware Depression Detection” and features two distinct tracks focused on young and elderly populations. This paper presents the dataset construction, personalized annotations, baseline models, and baeline results for both tracks. Our ablation studies demonstrate the effectiveness of personalized features in improving classification accuracy across multiple tasks. All code, datasets, and evaluation scripts are available through our official repository to support future research. We invite the community to participate in this challenge and contribute to advancing personalized mental health assessment.

\bibliographystyle{ACM-Reference-Format}
\bibliography{sample-base}


\begin{thebibliography}{18}


\ifx \showCODEN    \undefined \def \showCODEN     #1{\unskip}     \fi
\ifx \showISBNx    \undefined \def \showISBNx     #1{\unskip}     \fi
\ifx \showISBNxiii \undefined \def \showISBNxiii  #1{\unskip}     \fi
\ifx \showISSN     \undefined \def \showISSN      #1{\unskip}     \fi
\ifx \showLCCN     \undefined \def \showLCCN      #1{\unskip}     \fi
\ifx \shownote     \undefined \def \shownote      #1{#1}          \fi
\ifx \showarticletitle \undefined \def \showarticletitle #1{#1}   \fi
\ifx \showURL      \undefined \def \showURL       {\relax}        \fi
\providecommand\bibfield[2]{#2}
\providecommand\bibinfo[2]{#2}
\providecommand\natexlab[1]{#1}
\providecommand\showeprint[2][]{arXiv:#2}

\bibitem[Cai et~al\mbox{.}(2022)]%
        {cai2022multi}
\bibfield{author}{\bibinfo{person}{Hanshu Cai}, \bibinfo{person}{Zhenqin Yuan}, \bibinfo{person}{Yiwen Gao}, \bibinfo{person}{Shuting Sun}, \bibinfo{person}{Na Li}, \bibinfo{person}{Fuze Tian}, \bibinfo{person}{Han Xiao}, \bibinfo{person}{Jianxiu Li}, \bibinfo{person}{Zhengwu Yang}, \bibinfo{person}{Xiaowei Li}, {et~al\mbox{.}}} \bibinfo{year}{2022}\natexlab{}.
\newblock \showarticletitle{A multi-modal open dataset for mental-disorder analysis}.
\newblock \bibinfo{journal}{\emph{Scientific Data}} \bibinfo{volume}{9}, \bibinfo{number}{1} (\bibinfo{year}{2022}), \bibinfo{pages}{178}.
\newblock


\bibitem[Dibeklio{\u{g}}lu et~al\mbox{.}(2017)]%
        {dibekliouglu2017dynamic}
\bibfield{author}{\bibinfo{person}{Hamdi Dibeklio{\u{g}}lu}, \bibinfo{person}{Zakia Hammal}, {and} \bibinfo{person}{Jeffrey~F Cohn}.} \bibinfo{year}{2017}\natexlab{}.
\newblock \showarticletitle{Dynamic multimodal measurement of depression severity using deep autoencoding}.
\newblock \bibinfo{journal}{\emph{IEEE journal of biomedical and health informatics}} \bibinfo{volume}{22}, \bibinfo{number}{2} (\bibinfo{year}{2017}), \bibinfo{pages}{525--536}.
\newblock


\bibitem[Gratch et~al\mbox{.}(2014)]%
        {gratch2014distress}
\bibfield{author}{\bibinfo{person}{Jonathan Gratch}, \bibinfo{person}{Ron Artstein}, \bibinfo{person}{Gale~M Lucas}, \bibinfo{person}{Giota Stratou}, \bibinfo{person}{Stefan Scherer}, \bibinfo{person}{Angela Nazarian}, \bibinfo{person}{Rachel Wood}, \bibinfo{person}{Jill Boberg}, \bibinfo{person}{David DeVault}, \bibinfo{person}{Stacy Marsella}, {et~al\mbox{.}}} \bibinfo{year}{2014}\natexlab{}.
\newblock \showarticletitle{The distress analysis interview corpus of human and computer interviews.}. In \bibinfo{booktitle}{\emph{LREC}}. Reykjavik, \bibinfo{pages}{3123--3128}.
\newblock


\bibitem[Klein et~al\mbox{.}(2011)]%
        {klein2011personality}
\bibfield{author}{\bibinfo{person}{Daniel~N Klein}, \bibinfo{person}{Roman Kotov}, {and} \bibinfo{person}{Sara~J Bufferd}.} \bibinfo{year}{2011}\natexlab{}.
\newblock \showarticletitle{Personality and depression: explanatory models and review of the evidence}.
\newblock \bibinfo{journal}{\emph{Annual review of clinical psychology}} \bibinfo{volume}{7}, \bibinfo{number}{1} (\bibinfo{year}{2011}), \bibinfo{pages}{269--295}.
\newblock


\bibitem[Koelstra et~al\mbox{.}(2011)]%
        {koelstra2011deap}
\bibfield{author}{\bibinfo{person}{Sander Koelstra}, \bibinfo{person}{Christian Muhl}, \bibinfo{person}{Mohammad Soleymani}, \bibinfo{person}{Jong-Seok Lee}, \bibinfo{person}{Ashkan Yazdani}, \bibinfo{person}{Touradj Ebrahimi}, \bibinfo{person}{Thierry Pun}, \bibinfo{person}{Anton Nijholt}, {and} \bibinfo{person}{Ioannis Patras}.} \bibinfo{year}{2011}\natexlab{}.
\newblock \showarticletitle{Deap: A database for emotion analysis; using physiological signals}.
\newblock \bibinfo{journal}{\emph{IEEE transactions on affective computing}} \bibinfo{volume}{3}, \bibinfo{number}{1} (\bibinfo{year}{2011}), \bibinfo{pages}{18--31}.
\newblock


\bibitem[Lo et~al\mbox{.}(2017)]%
        {lo2017genome}
\bibfield{author}{\bibinfo{person}{Min-Tzu Lo}, \bibinfo{person}{David~A Hinds}, \bibinfo{person}{Joyce~Y Tung}, \bibinfo{person}{Carol Franz}, \bibinfo{person}{Chun-Chieh Fan}, \bibinfo{person}{Yunpeng Wang}, \bibinfo{person}{Olav~B Smeland}, \bibinfo{person}{Andrew Schork}, \bibinfo{person}{Dominic Holland}, \bibinfo{person}{Karolina Kauppi}, {et~al\mbox{.}}} \bibinfo{year}{2017}\natexlab{}.
\newblock \showarticletitle{Genome-wide analyses for personality traits identify six genomic loci and show correlations with psychiatric disorders}.
\newblock \bibinfo{journal}{\emph{Nature genetics}} \bibinfo{volume}{49}, \bibinfo{number}{1} (\bibinfo{year}{2017}), \bibinfo{pages}{152--156}.
\newblock


\bibitem[Miranda-Correa et~al\mbox{.}(2018)]%
        {miranda2018amigos}
\bibfield{author}{\bibinfo{person}{Juan~Abdon Miranda-Correa}, \bibinfo{person}{Mojtaba~Khomami Abadi}, \bibinfo{person}{Nicu Sebe}, {and} \bibinfo{person}{Ioannis Patras}.} \bibinfo{year}{2018}\natexlab{}.
\newblock \showarticletitle{Amigos: A dataset for affect, personality and mood research on individuals and groups}.
\newblock \bibinfo{journal}{\emph{IEEE transactions on affective computing}} \bibinfo{volume}{12}, \bibinfo{number}{2} (\bibinfo{year}{2018}), \bibinfo{pages}{479--493}.
\newblock


\bibitem[Moussavi et~al\mbox{.}(2007)]%
        {moussavi2007depression}
\bibfield{author}{\bibinfo{person}{Saba Moussavi}, \bibinfo{person}{Somnath Chatterji}, \bibinfo{person}{Emese Verdes}, \bibinfo{person}{Ajay Tandon}, \bibinfo{person}{Vikram Patel}, {and} \bibinfo{person}{Bedirhan Ustun}.} \bibinfo{year}{2007}\natexlab{}.
\newblock \showarticletitle{Depression, chronic diseases, and decrements in health: results from the World Health Surveys}.
\newblock \bibinfo{journal}{\emph{The Lancet}} \bibinfo{volume}{370}, \bibinfo{number}{9590} (\bibinfo{year}{2007}), \bibinfo{pages}{851--858}.
\newblock


\bibitem[Organization et~al\mbox{.}(2017)]%
        {world2017depression}
\bibfield{author}{\bibinfo{person}{World~Health Organization} {et~al\mbox{.}}} \bibinfo{year}{2017}\natexlab{}.
\newblock \showarticletitle{Depression and other common mental disorders: global health estimates}.
\newblock  (\bibinfo{year}{2017}).
\newblock


\bibitem[Ringeval et~al\mbox{.}(2019)]%
        {ringeval2019avec}
\bibfield{author}{\bibinfo{person}{Fabien Ringeval}, \bibinfo{person}{Bj{\"o}rn Schuller}, \bibinfo{person}{Michel Valstar}, \bibinfo{person}{Nicholas Cummins}, \bibinfo{person}{Roddy Cowie}, \bibinfo{person}{Leili Tavabi}, \bibinfo{person}{Maximilian Schmitt}, \bibinfo{person}{Sina Alisamir}, \bibinfo{person}{Shahin Amiriparian}, \bibinfo{person}{Eva-Maria Messner}, {et~al\mbox{.}}} \bibinfo{year}{2019}\natexlab{}.
\newblock \showarticletitle{AVEC 2019 workshop and challenge: state-of-mind, detecting depression with AI, and cross-cultural affect recognition}. In \bibinfo{booktitle}{\emph{Proceedings of the 9th International on Audio/visual Emotion Challenge and Workshop}}. \bibinfo{pages}{3--12}.
\newblock


\bibitem[Ringeval et~al\mbox{.}(2017)]%
        {ringeval2017avec}
\bibfield{author}{\bibinfo{person}{Fabien Ringeval}, \bibinfo{person}{Bj{\"o}rn Schuller}, \bibinfo{person}{Michel Valstar}, \bibinfo{person}{Jonathan Gratch}, \bibinfo{person}{Roddy Cowie}, \bibinfo{person}{Stefan Scherer}, \bibinfo{person}{Sharon Mozgai}, \bibinfo{person}{Nicholas Cummins}, \bibinfo{person}{Maximilian Schmitt}, {and} \bibinfo{person}{Maja Pantic}.} \bibinfo{year}{2017}\natexlab{}.
\newblock \showarticletitle{Avec 2017: Real-life depression, and affect recognition workshop and challenge}. In \bibinfo{booktitle}{\emph{Proceedings of the 7th annual workshop on audio/visual emotion challenge}}. \bibinfo{pages}{3--9}.
\newblock


\bibitem[Shen et~al\mbox{.}(2022)]%
        {shen2022automatic}
\bibfield{author}{\bibinfo{person}{Ying Shen}, \bibinfo{person}{Huiyu Yang}, {and} \bibinfo{person}{Lin Lin}.} \bibinfo{year}{2022}\natexlab{}.
\newblock \showarticletitle{Automatic depression detection: An emotional audio-textual corpus and a gru/bilstm-based model}. In \bibinfo{booktitle}{\emph{ICASSP 2022-2022 IEEE International Conference on Acoustics, Speech and Signal Processing (ICASSP)}}. IEEE, \bibinfo{pages}{6247--6251}.
\newblock


\bibitem[Valstar et~al\mbox{.}(2016)]%
        {valstar2016avec}
\bibfield{author}{\bibinfo{person}{Michel Valstar}, \bibinfo{person}{Jonathan Gratch}, \bibinfo{person}{Bj{\"o}rn Schuller}, \bibinfo{person}{Fabien Ringeval}, \bibinfo{person}{Denis Lalanne}, \bibinfo{person}{Mercedes Torres~Torres}, \bibinfo{person}{Stefan Scherer}, \bibinfo{person}{Giota Stratou}, \bibinfo{person}{Roddy Cowie}, {and} \bibinfo{person}{Maja Pantic}.} \bibinfo{year}{2016}\natexlab{}.
\newblock \showarticletitle{Avec 2016: Depression, mood, and emotion recognition workshop and challenge}. In \bibinfo{booktitle}{\emph{Proceedings of the 6th international workshop on audio/visual emotion challenge}}. \bibinfo{pages}{3--10}.
\newblock


\bibitem[Valstar et~al\mbox{.}(2014)]%
        {valstar2014avec}
\bibfield{author}{\bibinfo{person}{Michel Valstar}, \bibinfo{person}{Bj{\"o}rn Schuller}, \bibinfo{person}{Kirsty Smith}, \bibinfo{person}{Timur Almaev}, \bibinfo{person}{Florian Eyben}, \bibinfo{person}{Jarek Krajewski}, \bibinfo{person}{Roddy Cowie}, {and} \bibinfo{person}{Maja Pantic}.} \bibinfo{year}{2014}\natexlab{}.
\newblock \showarticletitle{Avec 2014: 3d dimensional affect and depression recognition challenge}. In \bibinfo{booktitle}{\emph{Proceedings of the 4th international workshop on audio/visual emotion challenge}}. \bibinfo{pages}{3--10}.
\newblock


\bibitem[Valstar et~al\mbox{.}(2013)]%
        {valstar2013avec}
\bibfield{author}{\bibinfo{person}{Michel Valstar}, \bibinfo{person}{Bj{\"o}rn Schuller}, \bibinfo{person}{Kirsty Smith}, \bibinfo{person}{Florian Eyben}, \bibinfo{person}{Bihan Jiang}, \bibinfo{person}{Sanjay Bilakhia}, \bibinfo{person}{Sebastian Schnieder}, \bibinfo{person}{Roddy Cowie}, {and} \bibinfo{person}{Maja Pantic}.} \bibinfo{year}{2013}\natexlab{}.
\newblock \showarticletitle{Avec 2013: the continuous audio/visual emotion and depression recognition challenge}. In \bibinfo{booktitle}{\emph{Proceedings of the 3rd ACM international workshop on Audio/visual emotion challenge}}. \bibinfo{pages}{3--10}.
\newblock


\bibitem[Yoon et~al\mbox{.}(2022)]%
        {yoon2022d}
\bibfield{author}{\bibinfo{person}{Jeewoo Yoon}, \bibinfo{person}{Chaewon Kang}, \bibinfo{person}{Seungbae Kim}, {and} \bibinfo{person}{Jinyoung Han}.} \bibinfo{year}{2022}\natexlab{}.
\newblock \showarticletitle{D-vlog: Multimodal vlog dataset for depression detection}. In \bibinfo{booktitle}{\emph{Proceedings of the AAAI Conference on Artificial Intelligence}}, Vol.~\bibinfo{volume}{36}. \bibinfo{pages}{12226--12234}.
\newblock


\bibitem[Zheng and Lu(2015)]%
        {zheng2015investigating}
\bibfield{author}{\bibinfo{person}{Wei-Long Zheng} {and} \bibinfo{person}{Bao-Liang Lu}.} \bibinfo{year}{2015}\natexlab{}.
\newblock \showarticletitle{Investigating critical frequency bands and channels for EEG-based emotion recognition with deep neural networks}.
\newblock \bibinfo{journal}{\emph{IEEE Transactions on autonomous mental development}} \bibinfo{volume}{7}, \bibinfo{number}{3} (\bibinfo{year}{2015}), \bibinfo{pages}{162--175}.
\newblock


\bibitem[Zou et~al\mbox{.}(2022)]%
        {zou2022semi}
\bibfield{author}{\bibinfo{person}{Bochao Zou}, \bibinfo{person}{Jiali Han}, \bibinfo{person}{Yingxue Wang}, \bibinfo{person}{Rui Liu}, \bibinfo{person}{Shenghui Zhao}, \bibinfo{person}{Lei Feng}, \bibinfo{person}{Xiangwen Lyu}, {and} \bibinfo{person}{Huimin Ma}.} \bibinfo{year}{2022}\natexlab{}.
\newblock \showarticletitle{Semi-structural interview-based Chinese multimodal depression corpus towards automatic preliminary screening of depressive disorders}.
\newblock \bibinfo{journal}{\emph{IEEE Transactions on Affective Computing}} \bibinfo{volume}{14}, \bibinfo{number}{4} (\bibinfo{year}{2022}), \bibinfo{pages}{2823--2838}.
\newblock


\end{thebibliography}










\end{document}